%% file: neurips_2024.tex
\documentclass{article}
\usepackage[preprint]{neurips_2024}

\usepackage[utf8]{inputenc} 
\usepackage[T1]{fontenc}    
\usepackage{hyperref}       
\usepackage{url}            
\usepackage{booktabs}       
\usepackage{amsfonts}       
\usepackage{nicefrac}       
\usepackage{microtype}      
\usepackage{xcolor}       
\usepackage{amsmath} 
\usepackage{graphicx}
\usepackage{algorithm}
\usepackage{algpseudocode}
\usepackage{subcaption}
\usepackage{tikz}
\usetikzlibrary{shapes.geometric, arrows.meta, positioning, calc}

\title{Reward-Based Online LLM Routing via NeuralUCB}

\author{
  Ming-Hua Tsai \\
  Oregon State University \\
  \texttt{tsaiming@oregonstate.edu}
  \And
  Phat Tran \\
  Oregon State University \\
  \texttt{tranphat@oregonstate.edu}
}

\begin{document}

\maketitle

\begin{abstract}
    This study investigates the use of NeuralUCB for cost-aware large language model (LLM) routing. Existing routing approaches can be broadly grouped into supervised routing methods and partial-feedback methods, each with different tradeoffs in efficiency and adaptivity. We implement a NeuralUCB-based routing policy and evaluate it on RouterBench under a simulated online setting. Experimental results show that the proposed method consistently outperforms random and min-cost baselines in utility reward. Compared with the max-quality reference, our method achieves substantially lower inference cost while maintaining competitive reward. These findings suggest that NeuralUCB is a promising approach for cost-aware LLM routing, while also highlighting remaining challenges in action discrimination and exploration.
\end{abstract}

\section{Problem Definition}
Large language models (LLMs) exhibit substantial variation in both capability and inference cost across tasks. High-end models often provide higher quality at higher cost, while cheaper models may fail on complex reasoning or coding queries. We study reward-based online LLM routing, which dynamically selects an appropriate LLM for each user query to balance answer quality and inference cost.

Formally, at interaction round $t$, the system observes a query $x_{t}$ and selects a model $a_{t}$ from a candidate set $\mathcal{A}=\{1,...,K\}$. To operationalize this in our framework, raw text queries are mapped into a dense semantic feature space using pre-trained sentence encoders. Since deployment typically invokes only one model per query, we only observe the feedback of the selected model $r_{t}=r(x_{t},a_{t})$ and do not observe outcomes for other models on the same query. The goal is to learn a routing policy that maximizes the long-horizon expected utility:
\[
\max_{\pi}\ \mathbb{E}\Big[\sum_{t=1}^{T} r(x_t,a_t)\Big],\qquad a_t\sim \pi(\cdot\mid x_t).
\]
A key challenge is that feedback is sparse and the mapping from queries to optimal model choices can be highly non-linear. We therefore cast routing as a contextual bandit problem and adopt NeuralUCB to learn a non-linear routing policy that improves average utility and the cost--quality tradeoff over time. 

\section{Related Work}
LLM routing aims to dynamically select an appropriate model for each query to balance answer quality and inference cost. A representative framework views routing as allocating queries of different difficulty to models with different cost profiles to reduce the average inference cost \cite{ding2024hybrid}. Existing approaches can be broadly categorized into two groups. Existing approaches can be broadly categorized into two groups. 

The first group is full-information supervised routing, which trains a router as a classification or ranking model using multiple candidate-model outputs and scores for the same query, such as RouterDC \cite{chen2024routerdc} and GraphRouter \cite{feng2024graphrouter}. While effective, these methods incur high data-collection cost and often require re-labeling and re-training when the query distribution or the model pool changes.

The second group learns from partial feedback and naturally fits the contextual bandit setting \cite{li2010contextual, zhou2015survey}. While LinUCB assumes linear rewards and kernelized methods (e.g., KernelUCB or GP-UCB) can be computationally expensive for high-dimensional embeddings, we adopt a neural contextual bandit perspective and use NeuralUCB to efficiently learn a non-linear query-to-model mapping under partial feedback while explicitly modeling the cost-quality tradeoff. To evaluate these approaches practically without the prohibitive cost of live API experimentation, this framework builds an offline replay pipeline over the RouterBench benchmark \cite{hu2024routerbench}. This allows for robust, split-level simulation of an online environment, enabling continuous tracking of cumulative rewards, action rates, and domain-specific performance (e.g., math versus coding).

\section{Methodology}
\subsection{Problem Setup and Utility Function}

We formulate cost-aware LLM routing as a contextual decision-making problem. For each input sample, we observe a context representation:
\[
x = (x_{\text{emb}}, x_{\text{feat}}, d),
\]
where $x_{\text{emb}}$ denotes the prompt embedding produced by a text encoder, $x_{\text{feat}}$ denotes auxiliary features, and $d$ denotes the domain of the sample. Given the context $x$, the policy selects an action:
\[
a \in \mathcal{A} = \{1,\dots,K\},
\]
from a set of candidate LLMs. In our setting, $K=11$.

For each sample-action pair, we consider two feedback signals, model quality and inference cost, denoted by:
\[
q(x,a), \qquad c(x,a).
\]
Under our simulated online LLM routing setting, the benchmark provides the quality and cost of all candidate models for each sample, which allows us to compare the utility of different actions over a fixed data stream.

Our objective is not to maximize model quality alone, but to achieve a tradeoff between quality and cost. To this end, we define the utility reward as:
\begin{equation}
\label{eq:utility_reward}
r(x,a)=q(x,a)\exp\bigl(-\lambda \tilde{c}(x,a)\bigr),
\end{equation}
where $\tilde{c}(x,a)$ is the normalized cost and $\lambda>0$ controls the strength of the cost penalty. To stabilize the reward scale, we apply logarithmic normalization to the raw cost:
\[
\tilde{c}(x,a)=\frac{\log\bigl(1+c(x,a)\bigr)}{\log\bigl(1+C_{\max}\bigr)},
\]
where $C_{\max}$ denotes the maximum observed cost in the dataset. This transformation maps $\tilde{c}(x,a)$ into $[0,1]$ and reduces instability caused by large cost-scale differences across candidate models.

\subsection{UtilityNet}
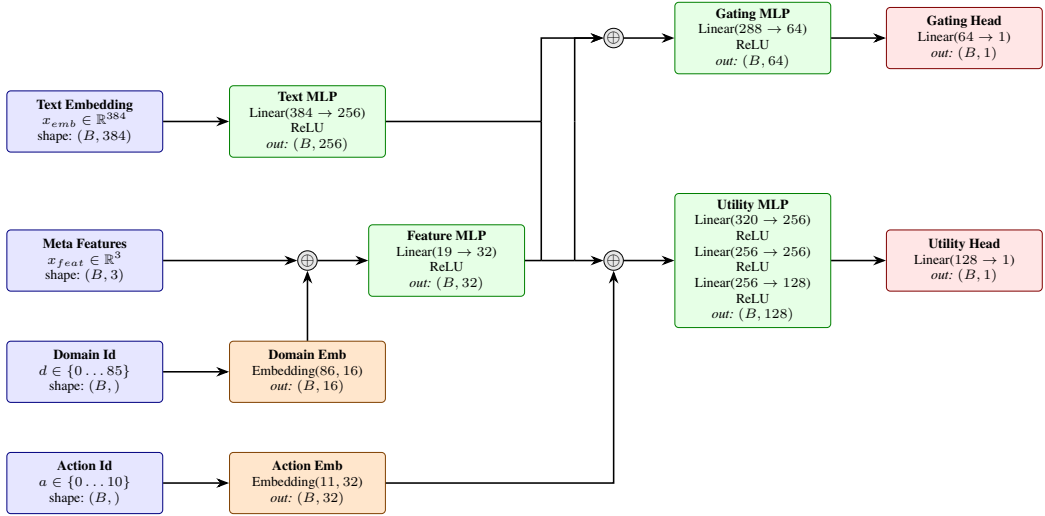
\begin{figure}[t]
    \centering
    \resizebox{\textwidth}{!}{
    \input{UtilityNet}
    }
    \caption{Architecture of UtilityNet. The utility branch predicts the utility reward for each context-action pair, while the gating branch predicts whether UCB-bonus-based action selection should be activated.}
    \label{fig:utilitynet}
\end{figure}

As shown in Figure \ref{fig:utilitynet}, UtilityNet serves as the utility predictor in our routing framework. Given a context-action pair, it estimates the corresponding utility reward, while its last hidden representation is further used by the NeuralUCB policy for uncertainty-aware action selection. We first encode the text input and auxiliary features as:
\[
h_{\text{emb}} = f_{\text{text}}(x_{\text{emb}}), \qquad
e_d = \mathrm{Emb}_d(d), \qquad
h_{\text{feat}} = f_{\text{feat}}([x_{\text{feat}}, e_d]),
\]
where $f_{\text{text}}(\cdot)$ and $f_{\text{feat}}(\cdot)$ denote the text and feature encoders, respectively, and $\mathrm{Emb}_d(\cdot)$ denotes the domain embedding.

For utility prediction, we define the action embedding:
\[
e_a = \mathrm{Emb}_a(a),
\]
and construct the fused representation:
\[
z_u(x,a) = [h_{\text{emb}}, h_{\text{feat}}, e_a].
\]
The utility branch then computes:
\[
h(x,a) = f_{\text{mlp}}(z_u(x,a)), \qquad
\mu(x,a) = f_{\text{u-head}}(h(x,a)),
\]
and is trained to regress Equation (\ref{eq:utility_reward}) using the Huber loss:
\[
\mathcal{L}_{u} = \mathrm{Huber}\bigl(\mu(x,a), r(x,a)\bigr).
\]

In addition, we include an auxiliary gating branch based on the context-dependent representation:
\[
z_g(x) = [h_{\text{emb}}, h_{\text{feat}}],
\]
which predicts:
\[
p(x) = f_{\text{g-head}}\bigl(f_{\text{gate}}(z_g(x))\bigr).
\]
The gating head is trained with a binary cross-entropy loss:
\[
\mathcal{L}_{g} = \mathrm{BCE}\bigl(p(x), y_{\text{gate}}(x)\bigr).
\]

\subsection{NeuralUCB-based Routing Policy}

Based on the utility prediction provided by UtilityNet, we further adopt NeuralUCB for action selection. Specifically, for each context-action pair, we use the last hidden representation $h(x,a)$ produced by UtilityNet as the feature representation for UCB, and define the augmented feature:
\[
g(x,a) = [h(x,a);1],
\]
where the appended constant $1$ serves as a bias term. Instead of maintaining separate uncertainty statistics for each action, we use a shared inverse covariance matrix:
\[
A^{-1},
\]
to estimate uncertainty in the last-layer feature space.

Under this setting, the UCB score of action $a$ is defined as:
\[
s(x,a)=\mu(x,a)+\beta\sqrt{g(x,a)^\top A^{-1}g(x,a)},
\]
where $\mu(x,a)$ is the utility score predicted by UtilityNet, and $\beta$ controls the strength of the exploration bonus. NeuralUCB therefore favors actions with both high predicted utility and high uncertainty.

In addition to the UCB score, we define a conservative fallback action:
\[
a^{\text{safe}}=\arg\max_{a\in\mathcal{A}}\mu(x,a),
\]
which selects the action solely based on the predicted mean utility. Using the gating probability $p(x)$ produced by the gating branch, the final action selection rule is:
\[
a^*=
\begin{cases}
\arg\max_{a\in\mathcal{A}} s(x,a), & p(x)\ge \tau_g,\\
a^{\text{safe}}, & \text{otherwise},
\end{cases}
\]
where $\tau_g$ denotes the gating threshold. In other words, the policy only activates UCB-bonus-based action selection when the gating branch predicts that exploration is beneficial. Otherwise, it falls back to the mean-based safe action.

We train the routing policy under a simulated online setting by splitting the data into 20 slices and processing them sequentially. For each slice, the policy selects actions, updates the replay buffer and shared $A^{-1}$, then trains UtilityNet and rebuilds $A^{-1}$ using the accumulated buffer data, as summarized in Algorithm \ref{alg:offline_neuralucb_split}.

\begin{algorithm}[t]
\caption{Simulated Online Protocol (Decide, Update, Train).}
\label{alg:offline_neuralucb_split}
\begin{algorithmic}[1]
\Require Slices $\{\mathcal{D}_t\}_{t=1}^{20}$, actions $\mathcal{A}=\{1,\ldots,11\}$, replay epochs $E=5$
\Require UtilityNet: mean $\mu_\theta(x,a)$, last hidden $h_\theta(x,a)$, gate prob $p_\theta(x)$
\Require Shared UCB stats: $A^{-1}$, hyperparams $\beta,\lambda,\tau_g$
\State Initialize buffer $\mathcal{B}\leftarrow\emptyset$
\State Initialize $A^{-1}\leftarrow \frac{1}{\lambda_0}I$
\For{$t=1$ to $20$}
    \For{each sample $i\in\mathcal{D}_t$}
        \State $a_i \leftarrow \textsc{Decide}(x_i;\theta,A^{-1},\beta,\tau_g)$
        \State $\textsc{Update}(\mathcal{B},A^{-1}; i,a_i,\theta)$
    \EndFor
    \State $\textsc{Train}(\theta;\mathcal{B},E,\lambda)$
    \State $A^{-1}\leftarrow \textsc{Rebuild}(A^{-1};\theta,\mathcal{B})$
\EndFor
\end{algorithmic}
\end{algorithm}

\section{Experimental}

\subsection{Experimental Setup}

We evaluate the proposed method on RouterBench, which contains 36,497 samples, 86 domains, and 11 candidate LLMs. For each sample, the benchmark provides the quality scores and inference costs of all candidate models. We adopt a simulated online protocol by splitting the data into 20 slices and processing them sequentially. After each slice, UtilityNet is trained on the replay buffer for 5 epochs, followed by rebuilding the shared $A^{-1}$.

We use all-MiniLM-L6-v2 \cite{wang2020minilm} as the text encoder, with learning rate $10^{-3}$, UCB bonus coefficient $\beta = 1$, and ridge regularization coefficient $\lambda = 1$. For comparison, we use three baselines: random, min-cost, and RouteLLM-BERT \cite{ong2024routellm}. For RouteLLM-BERT, we form a binary routing setup where the strong and weak models are defined as the models with the highest and lowest average utility reward, respectively.

\subsection{Experimental Results}
\begin{figure}[!htbp]
    \centering
    \begin{subfigure}[b]{0.495\textwidth}
        \centering
        \includegraphics[width=\linewidth]{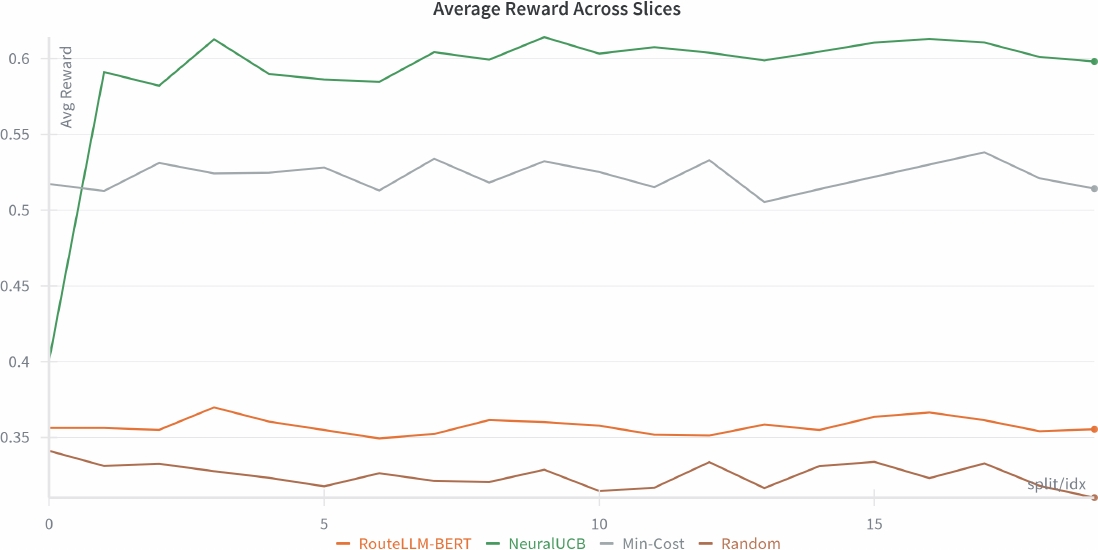}

        \caption{Average reward across slices.}
        \label{fig:avg_reward}
    \end{subfigure}
    \hfill
    \begin{subfigure}[b]{0.495\textwidth}
        \centering
        \includegraphics[width=\linewidth]{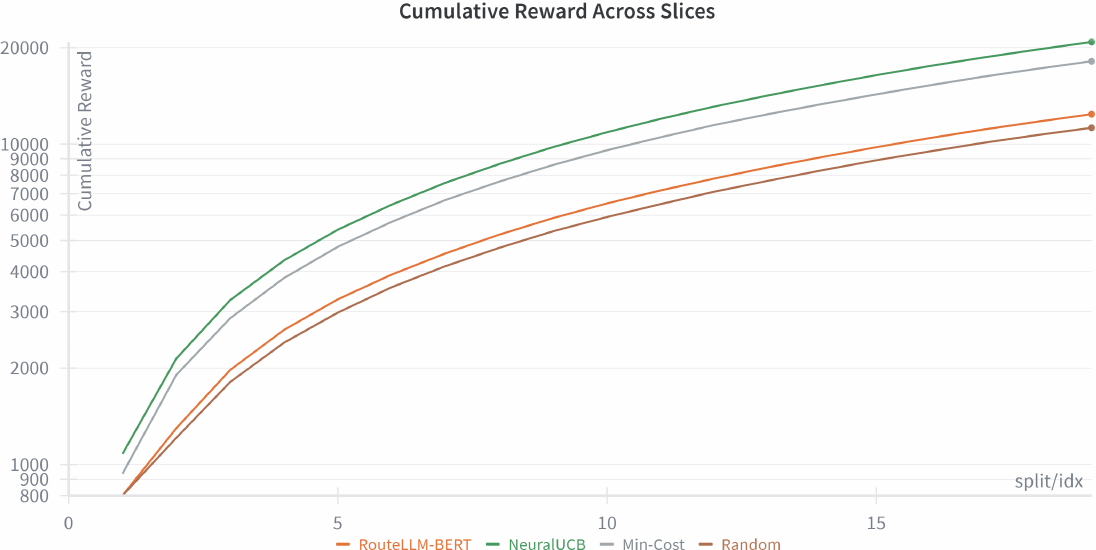}
        \caption{Cumulative reward across slices.}
        \label{fig:cum_reward}
    \end{subfigure}
    
    \caption{Reward comparison of NeuralUCB, RouteLLM-BERT, and two simple baselines, \textit{random} and \textit{min-cost}, under the simulated online routing setting.}
    \label{fig:combined_reward}
\end{figure}
Figure \ref{fig:avg_reward} and Figure \ref{fig:cum_reward} report the average reward and cumulative reward, respectively. From Figure \ref{fig:avg_reward}, NeuralUCB consistently outperforms RouteLLM-BERT as well as the \textit{random} and \textit{min-cost} baselines throughout the evaluation. In particular, the average reward of NeuralUCB remains relatively stable at around 0.59 to 0.61, whereas the \textit{min-cost} baseline stays around 0.51 to 0.53, RouteLLM-BERT remains around 0.35 to 0.36, and the \textit{random} baseline remains much lower, around 0.31 to 0.33. This indicates that the proposed policy learns a more favorable tradeoff between quality and cost than the compared baselines.

This trend becomes even more apparent in Figure \ref{fig:cum_reward}. As the number of processed slices increases, the cumulative reward of NeuralUCB remains consistently above RouteLLM-BERT and both simple baselines, and the gap continues to widen over time. The first slice is affected by the warm-start initialization and is therefore not used for formal comparison. Overall, these results show that the proposed method achieves more effective cost-aware routing than the baseline strategies.

\begin{figure}[!htbp]
    \centering
    \begin{subfigure}[b]{0.495\textwidth}
        \centering
        \includegraphics[width=\linewidth]{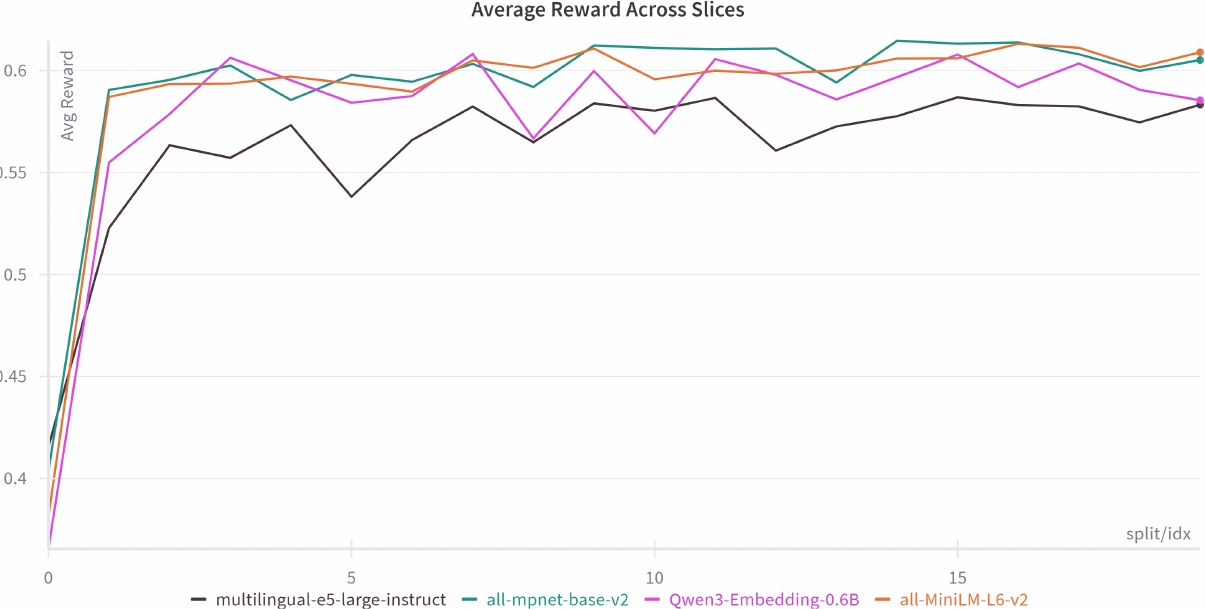}
        \caption{Average reward across slices.}
        \label{fig:textenc_avg_reward}
    \end{subfigure}
    \hfill
    \begin{subfigure}[b]{0.495\textwidth}
        \centering
        \includegraphics[width=\linewidth]{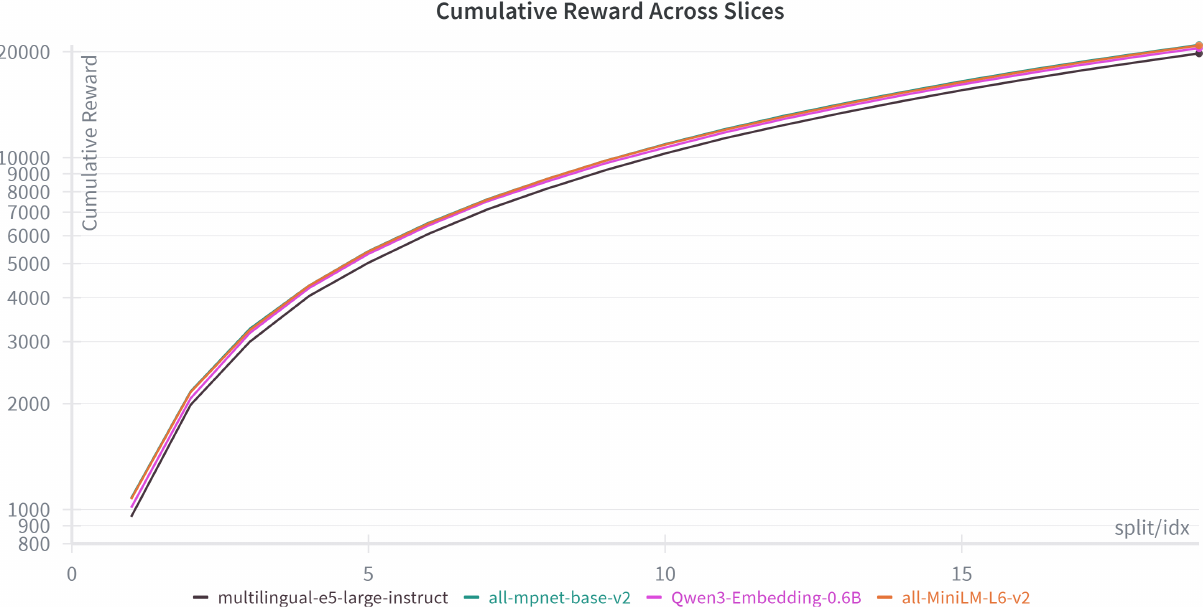}
        \caption{Cumulative reward across slices.}
        \label{fig:textenc_cum_reward}
    \end{subfigure}

    \caption{Encoder ablation under the simulated online routing setting. We compare four text encoders, \texttt{multilingual-e5-large-instruct}, \texttt{all-mpnet-base-v2}, \texttt{Qwen3-Embedding-0.6B}, and \texttt{all-MiniLM-L6-v2}, using both average reward and cumulative reward.}
    \label{fig:textenc_ablation}
\end{figure}

We compare four text encoders: \texttt{multilingual-e5-large-instruct} \cite{wang2024multilingual}, \texttt{all-mpnet-base-v2} \cite{reimers2019sentence,song2020mpnet}, \texttt{Qwen3-Embedding-0.6B} \cite{zhang2025qwen3}, and \texttt{all-MiniLM-L6-v2} \cite{wang2020minilm}.

Figure \ref{fig:textenc_avg_reward} and Figure \ref{fig:textenc_cum_reward} compare the effect of different text encoders on routing performance. Overall, \texttt{multilingual-e5-large-instruct} performs worst, whereas \texttt{all-mpnet-base-v2} and \texttt{all-MiniLM-L6-v2} achieve the strongest and most stable results. \texttt{Qwen3-Embedding-0.6B} remains competitive but exhibits slightly larger fluctuations. Apart from E5, the gap among the stronger encoders is relatively small, suggesting that changing to a larger encoder does not necessarily yield clear gains in this setting.

\begin{figure}[!htbp]
    \centering
    \begin{subfigure}[b]{0.495\textwidth}
        \centering
        \includegraphics[width=\linewidth]{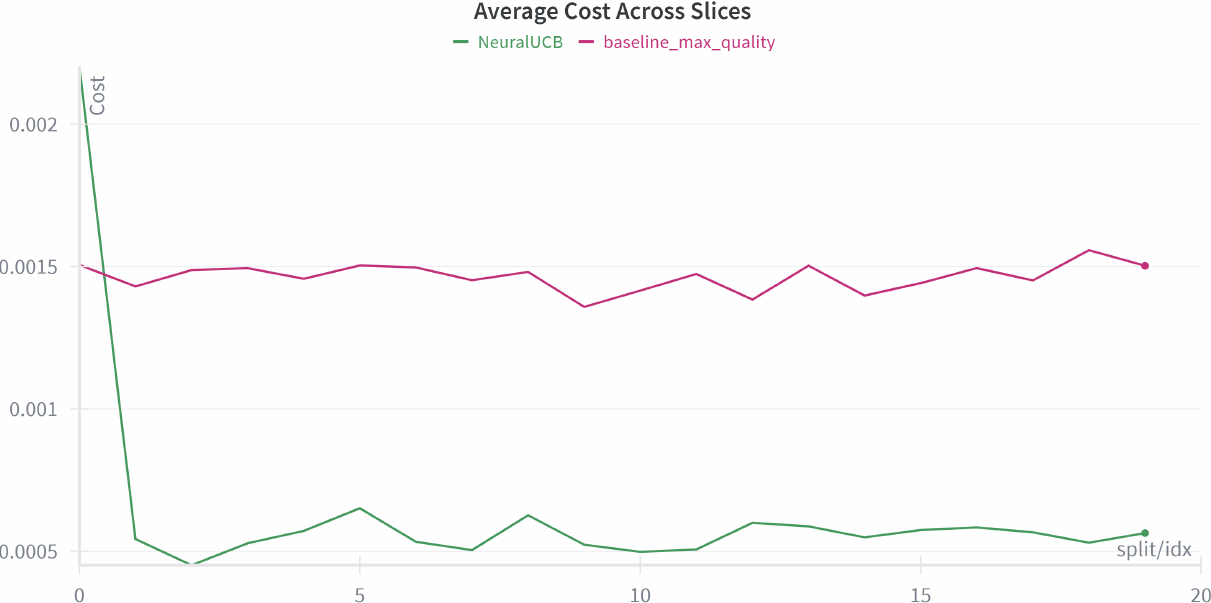}
        \caption{Average cost across slices.}
        \label{fig:maxq_cost}
    \end{subfigure}
    \hfill
    \begin{subfigure}[b]{0.495\textwidth}
        \centering
        \includegraphics[width=\linewidth]{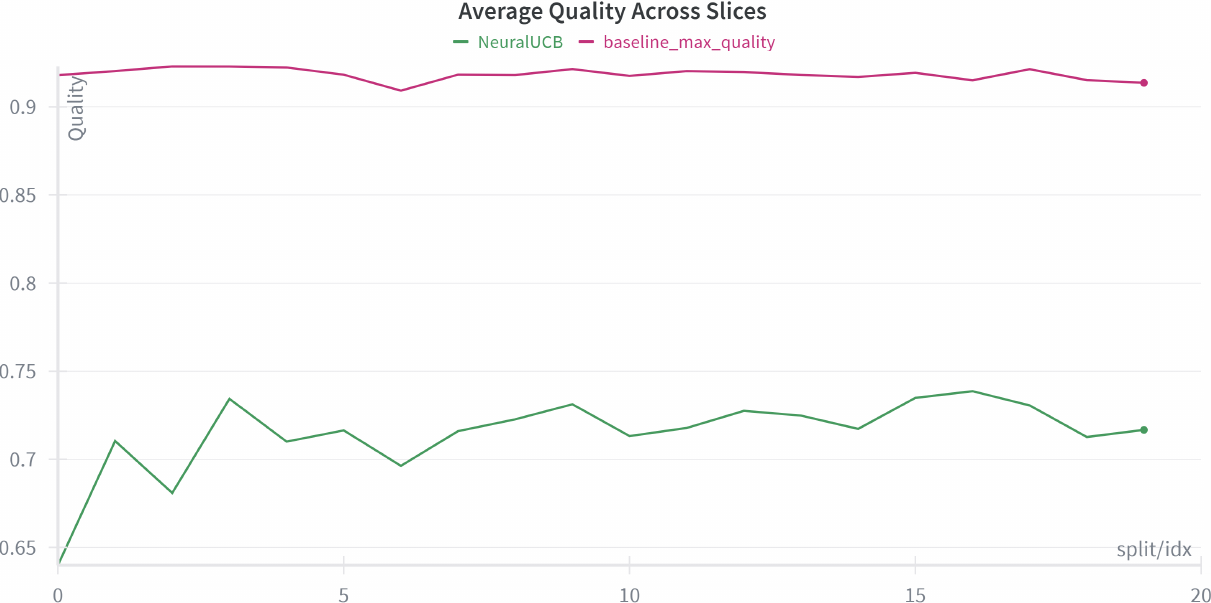}
        \caption{Average quality across slices.}
        \label{fig:maxq_quality}
    \end{subfigure}
    
    \caption{Comparison between the proposed NeuralUCB policy and the max-quality reference in terms of cost and selected quality under the simulated online routing setting.}
    \label{fig:maxq_tradeoff}
\end{figure}
As shown in Figure \ref{fig:maxq_cost} and Figure \ref{fig:maxq_quality}, the proposed NeuralUCB policy uses only about 33\% of the inference cost of the max-quality reference while maintaining competitive selected quality. Although a noticeable quality gap remains, the proposed policy still clearly outperforms the \textit{random} and \textit{min-cost} baselines in reward, indicating a meaningful cost-performance tradeoff. A remaining challenge is that, for many samples, the max-quality choice already aligns with the reward-optimal decision, limiting the potential gain from routing. Future work will therefore focus on collecting more training data and improving the utility model for better action discrimination under cost constraints.

\bibliographystyle{plainnat}
\bibliography{references}

\end{document}

%% file: UtilityNet.tex
\begin{tikzpicture}[
    base/.style={rectangle, draw, rounded corners=2pt, minimum width=2.8cm, minimum height=1.1cm, align=center, font=\scriptsize, inner sep=3pt},
    inputnode/.style={base, fill=blue!10, draw=blue!50!black},
    mlpnode/.style={base, fill=green!10, draw=green!50!black},
    embnode/.style={base, fill=orange!20, draw=orange!50!black},
    headnode/.style={base, fill=red!10, draw=red!50!black},
    sumjunction/.style={draw, circle, inner sep=0pt, minimum size=0.35cm, fill=white, font=\footnotesize},
    arrow/.style={-Stealth, thick}
]

\def\xColI{0}
\def\xColII{4}
\def\xColIII{6.5}
\def\xColIV{9.5} 
\def\xColV{12}
\def\xColVI{15.8}

\def\yRowGating{1.5} 
\def\yRowI{0}
\def\yRowII{-2.5}
\def\yRowIII{-4.5}
\def\yRowIV{-6.5}

\def\xColJuncGating{8.2}
\def\xColJuncUtility{8.8}

\node (text_emb) at (\xColI, \yRowI) [inputnode] {
    \textbf{Text Embedding}\\
    $x_{emb} \in \mathbb{R}^{384}$\\
    shape: $(B, 384)$
};
\node (meta_feat) at (\xColI, \yRowII) [inputnode] {
    \textbf{Meta Features}\\
    $x_{feat} \in \mathbb{R}^{3}$\\
    shape: $(B, 3)$
};
\node (domain_id) at (\xColI, \yRowIII) [inputnode] {
    \textbf{Domain Id}\\
    $d \in \{0\dots85\}$\\
    shape: $(B,)$
};
\node (action_id) at (\xColI, \yRowIV) [inputnode] {
    \textbf{Action Id}\\
    $a \in \{0\dots10\}$\\
    shape: $(B,)$
};

\node (sum_feat) at (\xColII, \yRowII) [sumjunction,  fill=gray!20] {$\oplus$};
\node (domain_emb) at (\xColII, \yRowIII) [embnode] {
    \textbf{Domain Emb}\\
    Embedding($86, 16$)\\
    \textit{out:} $(B, 16)$
};
\node (action_emb) at (\xColII, \yRowIV) [embnode] {
    \textbf{Action Emb}\\
    Embedding($11, 32$)\\
    \textit{out:} $(B, 32)$
};

\node (text_mlp) at (\xColII, \yRowI) [mlpnode] {
    \textbf{Text MLP}\\
    Linear($384 \to 256$)\\
    ReLU\\
    \textit{out:} $(B, 256)$
};
\node (feature_mlp) at (\xColIII, \yRowII) [mlpnode] {
    \textbf{Feature MLP}\\
    Linear($19 \to 32$)\\
    ReLU\\
    \textit{out:} $(B, 32)$
};

\node (sum_gating) at (\xColIV, \yRowGating) [sumjunction, fill=gray!20] {$\oplus$};
\node (sum_utility) at (\xColIV, \yRowII) [sumjunction, fill=gray!20] {$\oplus$};

\node (gating_mlp) at (\xColV, \yRowGating) [mlpnode] {
    \textbf{Gating MLP}\\
    Linear($288 \to 64$)\\
    ReLU\\
    \textit{out:} $(B, 64)$
};
\node (utility_mlp) at (\xColV, \yRowII) [mlpnode] {
    \textbf{Utility MLP}\\
    Linear($320 \to 256$)\\
    ReLU\\
    Linear($256 \to 256$)\\
    ReLU\\
    Linear($256 \to 128$)\\
    ReLU\\
    \textit{out:} $(B, 128)$
};

\node (gating_head) at (\xColVI, \yRowGating) [headnode] {
    \textbf{Gating Head}\\
    Linear($64 \to 1$)\\
    \textit{out:} $(B, 1)$
};
\node (utility_head) at (\xColVI, \yRowII) [headnode] {
    \textbf{Utility Head}\\
    Linear($128 \to 1$)\\
    \textit{out:} $(B, 1)$
};

\draw [arrow] (text_emb.east) -- (text_mlp.west);
\draw [arrow] (meta_feat.east) -- (sum_feat.west);
\draw [arrow] (domain_id.east) -- (domain_emb.west);
\draw [arrow] (action_id.east) -- (action_emb.west);
\draw [arrow] (sum_feat.east) -- (feature_mlp.west);
\draw [arrow] (sum_gating.east) -- (gating_mlp.west);
\draw [arrow] (sum_utility.east) -- (utility_mlp.west);
\draw [arrow] (gating_mlp.east) -- (gating_head.west);
\draw [arrow] (utility_mlp.east) -- (utility_head.west);

\draw [arrow] (domain_emb.north) -- (sum_feat.south);

\coordinate (JG_top) at (\xColJuncGating, \yRowI);
\coordinate (JU_top) at (\xColJuncUtility, \yRowI);
\coordinate (JG_bot) at (\xColJuncGating, \yRowII);
\coordinate (JU_bot) at (\xColJuncUtility, \yRowII);

\draw [thick] (text_mlp.east) -- (JG_top);
\draw [arrow] (JU_top) |- (sum_gating.west);
\draw [arrow] (JG_top) |- (sum_gating.west);
\draw [thick] (feature_mlp.east) -- (JG_bot);
\draw [thick] (JG_bot) -- (JU_bot);
\draw [arrow] (JU_bot) -- (sum_utility.west);
\draw [thick] (JG_top) -- (JG_bot); 
\draw [thick] (JU_top) -- (JU_bot); 
\draw [arrow] (action_emb.east) -| (sum_utility.south);

\end{tikzpicture}